%% file: main.tex
\documentclass[runningheads]{llncs}
\usepackage{etex}

\usepackage{times}
\usepackage{epsfig}
\usepackage{graphicx}
\usepackage{tabularx}
\usepackage{booktabs}
\usepackage{array}
\usepackage{color}
\usepackage[width=122mm,left=12mm,paperwidth=146mm,height=193mm,top=12mm,paperheight=217mm]{geometry}

\usepackage{amsmath}
\usepackage{bbm}
\usepackage{bm}
\usepackage{amssymb}
\usepackage{mathabx}
\usepackage[pagebackref=true,breaklinks=true,letterpaper=true,colorlinks,bookmarks=false]{hyperref}
\usepackage{lipsum}
\usepackage{multirow}
\usepackage{url}
\usepackage{lettrine}
\usepackage{cite}
\usepackage[ruled,vlined,linesnumbered]{algorithm2e}
\usepackage{pgfplots}
\usepackage{subcaption}
\usepackage{pgfplotstable}
\usepackage{pgf,tikz}
\usetikzlibrary{positioning}

\usepackage{appendix}

\newcommand{\CroW}{\textit{CroW}~} 
\newcommand{\Crow}{\CroW} 

\newcommand{\uCrow}{\textit{uCroW}~}
\newcommand{\sCrow}{\textit{SW}~} 
\newcommand{\cCrow}{\textit{SSW}~} 

\newcommand{\Sup}{\textit{\uCrow}}
\newcommand{\SuP}{\textit{\uCrow}}

\newcommand{\supp}{\textit{\CroW}}

\newcommand{\SCM}{\textit{\sCrow}}
\newcommand{\IDF}{\textit{\cCrow}}

\begin{document}
\pagestyle{headings}
\mainmatter

\title{Cross-dimensional Weighting for \\ Aggregated Deep Convolutional Features}

\authorrunning{Kalantidis, Mellina \& Osindero}
\author{Yannis Kalantidis, Clayton Mellina \and Simon Osindero}
\institute{Computer Vision and Machine Learning Group \\ Flickr, Yahoo\\
}

\maketitle

\input{tex/abbrev}
\input{tex/plots}

\input{tex/abstract}
\input{tex/intro}
\input{tex/related}

\input{tex/framework}

\input{tex/feature_weighting}
\input{tex/experiments}

\input{tex/conclusions}

\bibliographystyle{splncs03}
\bibliography{tex/bib}

\end{document}

%% file: tex/abbrev.tex
\newcommand{\nospace}{\vspace{-4mm}}
\newcommand{\head}[1]{{\smallskip\noindent\textbf{#1}}}
\newcommand{\alert}[1]{{\color{red}{#1}}}
\newcommand{\suppl}{\emph{given in the supplementary material}}


\newcommand{\one}{\mathbbm{1}}
\newcommand{\bbF}{\mathbb{F}}
\newcommand{\bbH}{\mathbb{H}}
\newcommand{\expect}{\mathbb{E}}
\newcommand{\project}{\mathbb{P}}
\newcommand{\real}{\mathbb{R}}
\newcommand{\tran}{^{\mathrm{T}}}
\newcommand{\diff}{\mathrm{d}}
\newcommand{\prob}{\mathrm{Pr}}
\newcommand{\binomial}{\mathrm{Bi}}
\newcommand{\weibull}{\mathrm{Wb}}
\newcommand{\normal}{\mathcal{N}}
\newcommand{\wishart}{\mathcal{W}}
\newcommand{\zcol}{\mathbf{0}}
\newcommand{\zrow}{\zcol\tran}
\newcommand{\otherwise}{\mathrm{otherwise}}

\newcommand{\cA}{\mathcal{A}}
\newcommand{\cB}{\mathcal{B}}
\newcommand{\cC}{\mathcal{C}}
\newcommand{\cD}{\mathcal{D}}
\newcommand{\cE}{\mathcal{E}}
\newcommand{\cF}{\mathcal{F}}
\newcommand{\cG}{\mathcal{G}}
\newcommand{\cH}{\mathcal{H}}
\newcommand{\cI}{\mathcal{I}}
\newcommand{\cJ}{\mathcal{J}}
\newcommand{\cK}{\mathcal{K}}
\newcommand{\cL}{\mathcal{L}}
\newcommand{\cM}{\mathcal{M}}
\newcommand{\cN}{\mathcal{N}}
\newcommand{\cO}{\mathcal{O}}
\newcommand{\cP}{\mathcal{P}}
\newcommand{\cQ}{\mathcal{Q}}
\newcommand{\cR}{\mathcal{R}}
\newcommand{\cS}{\mathcal{S}}
\newcommand{\cT}{\mathcal{T}}
\newcommand{\cU}{\mathcal{U}}
\newcommand{\cV}{\mathcal{V}}
\newcommand{\cW}{\mathcal{W}}
\newcommand{\cX}{\mathcal{X}}
\newcommand{\cY}{\mathcal{Y}}
\newcommand{\cZ}{\mathcal{Z}}

\newcommand{\cII}{\mathcal{I}_f}
\newcommand{\cAA}{\mathcal{A}_f}
\newcommand{\It}{i}
\newcommand{\At}{a}

\newcommand{\va}{\mathbf{a}}
\newcommand{\vb}{\mathbf{b}}
\newcommand{\vc}{\mathbf{c}}
\newcommand{\vd}{\mathbf{d}}
\newcommand{\ve}{\mathbf{e}}
\newcommand{\vf}{\mathbf{f}}
\newcommand{\vg}{\mathbf{g}}
\newcommand{\vh}{\mathbf{h}}
\newcommand{\vi}{\mathbf{i}}
\newcommand{\vj}{\mathbf{j}}
\newcommand{\vk}{\mathbf{k}}
\newcommand{\vl}{\mathbf{l}}
\newcommand{\vm}{\mathbf{m}}
\newcommand{\vn}{\mathbf{n}}
\newcommand{\vo}{\mathbf{o}}
\newcommand{\vp}{\mathbf{p}}
\newcommand{\vq}{\mathbf{q}}
\newcommand{\vr}{\mathbf{r}}
\newcommand{\vt}{\mathbf{t}}
\newcommand{\vu}{\mathbf{u}}
\newcommand{\vv}{\mathbf{v}}
\newcommand{\vw}{\mathbf{w}}
\newcommand{\vx}{\mathbf{x}}
\newcommand{\vy}{\mathbf{y}}
\newcommand{\vz}{\mathbf{z}}

\newcommand{\vA}{\mathbf{A}}
\newcommand{\vB}{\mathbf{B}}
\newcommand{\vC}{\mathbf{C}}
\newcommand{\vD}{\mathbf{D}}
\newcommand{\vF}{\mathbf{F}}
\newcommand{\vG}{\mathbf{G}}
\newcommand{\vI}{\mathbf{I}}
\newcommand{\vM}{\mathbf{M}}
\newcommand{\vS}{\mathbf{S}}
\newcommand{\vW}{\mathbf{W}}
\newcommand{\vX}{\mathbf{X}}
\newcommand{\vzero}{\mathbf{0}}

\newcommand{\rLambda}{\mathrm{\Lambda}}
\newcommand{\rSigma}{\mathrm{\Sigma}}

\newcommand{\vmu}{\bm{\mu}}
\newcommand{\vpi}{\bm{\pi}}
\newcommand{\vLambda}{\bm{\rLambda}}
\newcommand{\vSigma}{\bm{\rSigma}}


\newcommand{\rep}[2]{%
   \ifthenelse{\equal{\version}{track}}%
      {{\color{gray}
{#1}}{\color{blue}{#2}}}
      {\ifthenelse{\equal{\version}{long}}{#1}{#2}}%
}

\newcommand{\cut}[1]{\rep{#1}{}}
\newcommand{\ins}[1]{\rep{}{#1}}


%
%
%

\def\onedot{.~}

\def\eg{\emph{e.g}.,~} \def\Eg{\emph{E.g}\onedot}
\def\ie{\emph{i.e}.,~} \def\Ie{\emph{I.e}\onedot}
\def\cf{\emph{c.f}\onedot} \def\Cf{\emph{C.f}\onedot}
\def\etc{\emph{etc}\onedot} 
\def\wrt{w.r.t\onedot} \def\dof{d.o.f\onedot}
\def\etal{\emph{et al}\onedot}
\makeatother

\newcommand{\boldchannelsAtLocation}{\bm{\lambda}^{(ij)}}
\newcommand{\boldmatrixForChannel}{\bm{\cC}^{(k)}}
\newcommand{\channelsAtLocation}{\lambda^{(ij)}}
\newcommand{\matrixForChannel}{\cC^{(k)}}

\newcommand{\spatialWeight}{\alpha_{ij}}
\newcommand{\channelWeight}{\beta_{k}}

%% file: tex/plots.tex

\newcommand{\plot}[2]{\cffolderinput{#1}{#2}}
\newcommand{\data}[1]{\cfcurrentfolder #1.txt}
\newcommand{\tabread}[2]{\pgfplotstableread{\data{#1}}#2}

\newcommand{\leg}[1]{\addlegendentry{#1}}
\newcommand{\legtext}[1]{\addlegendimage{empty legend} \addlegendentry{#1}}

\tikzset{every mark/.append style={solid}}
\pgfplotsset{ grid=both, width=\columnwidth, try min ticks=5,
	 every axis x label/.style={at={(ticklabel cs:0.5)},anchor=north},
	 every axis y label/.style={at={(ticklabel cs:0.5)},rotate=90,anchor=south},
	 every axis/.append style={font=\scriptsize,thick,mark=*,smooth,tension=0.18},
	 legend cell align=left, legend style={fill opacity=0.7}
}

\pgfplotsset{
    dash/.style={mark=o,dashed,opacity=0.6},
    mmark/.style={mark=*},
}
\newcommand{\kilo}[1]{\thisrow{#1}/1000}


%% file: tex/abstract.tex
\begin{abstract}

\noindent
We propose a simple and straightforward way of creating powerful image representations via cross-dimensional weighting and aggregation of deep convolutional neural network layer outputs. We first present a generalized framework that encompasses a broad family of approaches and includes cross-dimensional pooling and weighting steps. We then propose specific non-parametric schemes for both spatial- and channel-wise weighting that boost the effect of highly active spatial responses and at the same time regulate burstiness effects. We experiment on different public datasets for image search and show that our approach outperforms the current state-of-the-art for approaches based on pre-trained networks. We also provide an easy-to-use, open source implementation  that reproduces our results.
\end{abstract}

%% file: tex/intro.tex
\section{Introduction}
\label{sec:intro}

\noindent
Visual image search has been evolving rapidly in recent years with hand-crafted local features giving way to learning-based ones. Deep Convolutional Neural Networks (CNNs) were popularized by the seminal work of Krizhevsky~\etal~\cite{KrSH12} and have been shown to ``effortlessly'' improve the state-of-the-art in multiple computer vision domains~\cite{RASC14}, beating many highly optimized, domain-specific approaches.
It comes as no surprise that such features, based on deep networks, have recently also dominated the field of visual image search~\cite{RASC14, BSCL14,ARS+14, BaLe15}. 

Many recent image search approaches are based on deep features, \eg Babenko~\etal~\cite{BSCL14,BaLe15} and Razavian~\etal~\cite{RASC14,ARS+14} proposed different pooling strategies for such features and demonstrated state-of-the-art performance in popular benchmarks for \textit{compact} image representations, \ie representations of up to a few hundred dimensions.

Motivated by these advances, in this paper we present a simple and straightforward way of creating powerful image representations via cross-dimensional weighting and aggregation. We place our approach in a general family of approaches for multidimensional aggregation and weighting and present a specific instantiation that we have thus far found to be most effective on benchmark tasks.

We base our cross-dimensional weighted features on a generic deep convolutional neural network. Since we aggregate outputs of convolutional layers before the fully connected ones, the data layer can be of arbitrary size~\cite{LoSD14}. We therefore avoid resizing and cropping the input image, allowing images of different aspect ratios to keep their spatial characteristics intact. After extracting deep convolutional features from the last spatial layer of a CNN, we apply weighting both spatially and per channel before sum-pooling to create a final aggregation. We denote features derived after such cross-dimensional weighting and pooling as \CroW features.


Our contributions can be summarized as follows:
\begin{itemize}
\setlength\itemsep{.01em}
\item We present a generalized framework that sketches a family of approaches for aggregation of convolutional features, including cross-dimensional weighting and pooling steps.
\item We propose non-parametric weighting schemes for both spatial- and channel-wise weighting that boost the effect of highly active spatial responses and regulate the effect of channel burstiness respectively. 
\item We present state-of-the-art results on three public datasets for image search without any fine-tuning. 
\end{itemize}

With a very small computational overhead, we are able to improve the state-of-the-art in visual image search.
For the popular \emph{Oxford}~\cite{PCI+07} and \emph{Paris}~\cite{PCS+08} datasets, the mean average precision for our \Crow feature is over $10\%$ higher than the previous state-of-the-art for compact visual representations. Additionally, our features are trivially combined for simple query expansion, enjoying even better performance.
\textit{We provide an easy-to-use, open source implementation that reproduces our results on GitHub}\footnote{\url{https://github.com/yahoo/crow}}.


The paper is structured as follows: In Section~\ref{sec:related} we present and discuss related work, while in Section~\ref{sec:framework} we present a general framework for weighted pooling to orient past work and our own explorations. In Section~\ref{sec:weighting} we describe two complimentary feature weighting schemes, and we present experimental results for visual search in Section~\ref{sec:exp}. The paper concludes with Section~\ref{sec:conclusions}.

%% file: tex/related.tex
\section{Related work}
\label{sec:related}

Until recently, the vast majority of image search approaches were variants of the bag-of-words model~\cite{SiZi03} and were based on local features, typically SIFT~\cite{Lowe01}. 
Successful extensions include soft assignment~\cite{PCS+08}, spatial matching~\cite{PCI+07,AvTo13}, query expansion~\cite{CPS+07, CMPM11, ArZi12, ToJe14}, better descriptor normalization~\cite{ArZi12}, feature selection~\cite{TuLo09,TKAK14}, feature burstiness~\cite{JeDS09} and very large vocabularies~\cite{MPCM10}. 
All the aforementioned strategies perform very well for object retrieval but are  very hard to scale, as each image is represented by hundreds of patches, causing search time and memory to suffer.

The community therefore recently turned towards global image representations. Starting from local feature aggregation strategies like VLAD~\cite{JDSP10} or Fisher Vectors~\cite{PLSP10} multiple successful extensions have arisen
~\cite{ToAJ13,DGJP13,GMJP14,ToJA15},
slowly increasing the performance of such aggregated features and closing the gap between global and bag-of-word representations for image search. Triangulation embedding with democratic aggregation~\cite{JeZi14} was shown to give state-of-the-art results for SIFT-based architectures, while handling problems related to burstiness and interactions between unrelated descriptors prior to aggregation. Recently, Murray and Perronnin~\cite{MuPe14} generalized max-pooling from bag-of-words to Fisher Vector representations achieving high performance in search as well as classification tasks.

After the seminal work of Krizhevsky~\etal~\cite{KrSH12}, image search, along with the whole computer vision community, embraced the power of deep learning architectures. Out-of-the-box features from pre-trained Convolutional Neural Networks (CNNs) were shown to effortlessly give state-of-the-art results in many computer vision tasks, including image search~\cite{RASC14}.

Among the first to more extensively study CNN-based codes for image search were Babenko~\etal~\cite{BSCL14} and Razavian~\etal~\cite{RASC14, ARS+14}. They experimented with aggregation of responses from different layers of the CNN, both fully connected and convolutional. They introduced a basic feature aggregation pipeline using max-pooling that, in combination with proper normalization and whitening was able to beat all aggregated local feature based approaches for low dimensional image codes. Gong~\etal~\cite{GWGL14} used orderless VLAD pooling of CNN activations on multiple scales and achieved competitive results on classification and search tasks.

Very recently Tolias~\etal~\cite{ToSJ15} proposed max-pooling over multiple image regions sampled on the final convolutional layer. Their approach achieves state-of-the-art results and is complementary to our cross-dimensional weighting.
Cimpoi~\etal~\cite{CiMV15} also recently proposed using Fisher Vector aggregation of convolutional features for texture recognition. Their approach achieves great performace, it is however computationally demanding; PCA from 65K dimensions alone requires multiplication with a very large matrix. Our approach is training- and parameter- free, with only a very small computational overhead.

In another very recent related work, Babenko and Lempitsky proposed the SPoC features~\cite{BaLe15} with slightly different design choices from the pipeline of~\cite{BSCL14} and sum- instead of max-pooling. As the latter approach is very related to ours, we discuss the differences of the two approaches in the following sections and explain SPoC in terms of the proposed aggregation framework. 

The first approaches that \textit{learn} features for landmark retrieval~\cite{RaTC16, GARL16} are presented at the current ECCV conference. Both approaches use clean annotated data and fine-tune a deep CNN for feature extraction using a pairwise~\cite{RaTC16} or ranking~\cite{GARL16} loss. These approaches are now state-of-the art in the most common benchmarks. Still, our proposed features are not far behind, without requiring training or clean annotated data.



%% file: tex/framework.tex
\section{Framework for Aggregation of Convolutional Features}
\label{sec:framework}

\subsection{Framework Overview}

In this section we present a simple and straightforward way of creating powerful image representations.
We start by considering a general family of approaches that can be summarized as proceeding through the following steps. Greater details and motivations for these steps will be given in subsequent sections, along with the specific instantiation that we have thus far found to be most effective on benchmark tasks. 



\begin{description}
  \item[1: Perform spatially-local pooling.]
	Sum-pooling or max-pooling over a spatially local neighborhood within each channel of a convolutional layer, with neighborhood size $w{\times}h$ and stride $s$. Some limiting cases include: (1) a pooling neighborhood that occupies the full spatial extent of each channel (i.e. global pooling); and (2) a $1{\times}1$ pooling neighborhood (effectively not doing pooling at all). After pooling, we have a three-dimensional tensor of activities.
  \item[2: Compute spatial weighting factors.]
  For each location $(i,j)$ in the locally pooled feature maps we assign a weight, $\spatialWeight$, that is applied to each channel at that location.
  \item[3: Compute channel weighting factors.]
  For each channel $k$, we assign a weight, $\channelWeight$ that is applied to each location in that channel.
  \item[4: Perform weighted-sum aggregation.]
  We apply the previously derived weights location-wise and channel-wise before using a channel-wise sum to aggregate the full tensor of activities into a single vector.
  \item[5: Perform vector normalization.]
 The resulting vector is then normalized and power-transformed. A variety of norms can be used here.
  \item[6: Perform dimensionality reduction.]
 We then reduce the dimensionality of the normed-vector. PCA is a typical choice here, and we may also choose to perform whitening or other per-dimension scalings on entries of the dimensionality reduced vector.
   \item[7: Perform final normalization.]
 We then apply a second and final normalization step.
\end{description}

Algorithm~\ref{alg:general_features} summarises these steps as pseudocode. 

\input{fig/feature_weighting}

\subsection{Cross-dimensional Weighting}

Let $\bm{\cX} \in \mathbb{R}^{(K \times W \times H)}$ be the 3-dimensional \textit{feature} tensor from a selected layer $l$, where $K$ is the total number of channels and $W$, $H$ the spatial dimensions of that layer. As mentioned above, the spatial dimensions may vary per image depending on its original size, but we omit image-specific subscripts here for clarity.

We denote the entry in $\bm{\cX}$ corresponding to channel $k$, at spatial location $(i,j)$ as $\cX_{kij}$. For notational convenience, we also denote the channel-wise matrices of $\bm{\cX}$ as $\boldmatrixForChannel$, where $\matrixForChannel_{ij} = \cX_{kij}$. Similarly, we use $\boldchannelsAtLocation$ to denote the vector of channel responses at location $(i,j)$, where $\channelsAtLocation_k = \cX_{kij}$.

A weighted feature tensor $\bm{\cX^\prime}$ is produced by applying per-location weights, $\spatialWeight$, and per-channel weights, $\channelWeight$, to feature tensor $\bm{\cX}$ as illustrated in Figure~\ref{fig:feature_weighting}:

\begin{equation}
\cX_{kij}^\prime = \spatialWeight \channelWeight \cX_{kij}
\label{eqn:spatialweightsum}
\end{equation}

The weighted feature tensor is aggregated by sum-pooling per channel. Let \textit{aggregated feature} vector $\bm{\cF} = \{f_1, \ldots, f_k\}$ associated with the layer $l$ be the vector of weight-summed activations per channel:

\begin{equation}
f_k = {\sum\limits_{i=1}^{W}}{\sum\limits_{j=1}^{H} \cX_{kij}^\prime}
\end{equation}

After aggregation, we follow what was shown to be the best practice~\cite{RASC14, ARS+14} and L2-normalize $\bm{\cF}$, then whiten using parameters learnt from a separate dataset and L2-normalize again.
We denote the features that are derived from the current framework as \textit{Cross-dimensional Weighted} or \CroW features.

\input{tex/algorithms_v2}


%% file: fig/feature_weighting.tex
\begin{figure}[t]
\centering
\begin{tikzpicture}

	\node(raw_feature){
		\begin{tikzpicture}
			\pgfmathsetmacro{\cubex}{0}
	    	\pgfmathsetmacro{\cubey}{0}
			\pgfmathsetmacro{\cubew}{2}
			\pgfmathsetmacro{\cubeh}{1}
			\pgfmathsetmacro{\cubed}{1}
			\draw[black] (0,0,0) -- ++(-\cubew,0,0) -- ++(0,-\cubeh,0) -- ++(\cubew,0,0) -- cycle;
			\draw[black] (0,0,0) -- ++(0,0,-\cubed) -- ++(0,-\cubeh,0) -- ++(0,0,\cubed) -- cycle;
			\draw[black] (0,0,0) -- ++(-\cubew,0,0) -- ++(0,0,-\cubed) -- ++(\cubew,0,0) -- cycle;
		\end{tikzpicture}
	};
    \node(raw_feature_label)[below=-0.8cm of raw_feature]{$\cX$};
    
    \node[right=0.1 of raw_feature](spatial_weight){
		\begin{tikzpicture}
			\pgfmathsetmacro{\cubex}{0}
	    	\pgfmathsetmacro{\cubey}{0}
			\pgfmathsetmacro{\cubew}{0.2}
			\pgfmathsetmacro{\cubeh}{1}
			\pgfmathsetmacro{\cubed}{1}
			\draw[black] (0,0,0) -- ++(-\cubew,0,0) -- ++(0,-\cubeh,0) -- ++(\cubew,0,0) -- cycle;
			\draw[black] (0,0,0) -- ++(0,0,-\cubed) -- ++(0,-\cubeh,0) -- ++(0,0,\cubed) -- cycle;
			\draw[black] (0,0,0) -- ++(-\cubew,0,0) -- ++(0,0,-\cubed) -- ++(\cubew,0,0) -- cycle;
		\end{tikzpicture}
	};
    \node(spatial_weight_label)[right=.5mm of spatial_weight]{$\alpha$};
    
    \node[below=.1cm of raw_feature](channel_weight){
		\begin{tikzpicture}
			\pgfmathsetmacro{\cubex}{0}
	    	\pgfmathsetmacro{\cubey}{0}
			\pgfmathsetmacro{\cubew}{2}
			\pgfmathsetmacro{\cubeh}{0.2}
			\pgfmathsetmacro{\cubed}{0.2}
			\draw[black] (0,0,0) -- ++(-\cubew,0,0) -- ++(0,-\cubeh,0) -- ++(\cubew,0,0) -- cycle;
			\draw[black] (0,0,0) -- ++(0,0,-\cubed) -- ++(0,-\cubeh,0) -- ++(0,0,\cubed) -- cycle;
			\draw[black] (0,0,0) -- ++(-\cubew,0,0) -- ++(0,0,-\cubed) -- ++(\cubew,0,0) -- cycle;
		\end{tikzpicture}
	};
    \node(channel_weight_label)[below=.1cm of channel_weight]{$\beta$};

	\node(weighted_feature)[right=2.8 of raw_feature]{
		\begin{tikzpicture}
			\pgfmathsetmacro{\cubex}{0}
	    	\pgfmathsetmacro{\cubey}{0}
			\pgfmathsetmacro{\cubew}{2}
			\pgfmathsetmacro{\cubeh}{1}
			\pgfmathsetmacro{\cubed}{1}
			\draw[black] (0,0,0) -- ++(-\cubew,0,0) -- ++(0,-\cubeh,0) -- ++(\cubew,0,0) -- cycle;
			\draw[black] (0,0,0) -- ++(0,0,-\cubed) -- ++(0,-\cubeh,0) -- ++(0,0,\cubed) -- cycle;
			\draw[black] (0,0,0) -- ++(-\cubew,0,0) -- ++(0,0,-\cubed) -- ++(\cubew,0,0) -- cycle;
		\end{tikzpicture}
	};
    \node(weighted_feature_label)[below=-0.8cm of weighted_feature]{$\cX^\prime$};
    
    \draw[ultra thick, ->] (spatial_weight_label) -- (weighted_feature);

\end{tikzpicture}

\caption{Prior to aggregation, the convolutional features can be weighted channel-wise by a weight vector $\beta$ and weighted location-wise by a weight matrix $\alpha$ such that $\cX_{kij}^\prime = \spatialWeight \channelWeight \cX_{kij}$. The weighted features $\cX^\prime$ are sum-pooled to derive an aggregate feature.}
\label{fig:feature_weighting}
\end{figure}
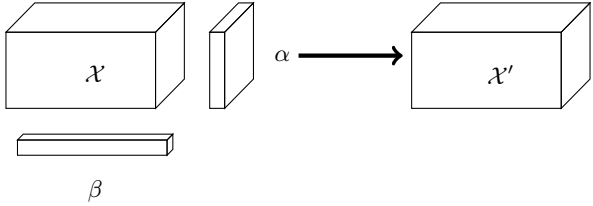

%% file: tex/algorithms_v2.tex
\begin{algorithm}[t]

\DontPrintSemicolon
\SetFuncSty{textsc}
\SetDataSty{emph}
\newcommand{\commentsty}[1]{{\color{green!50!black}#1}}
\SetCommentSty{commentsty}

\SetKwInOut{Input}{input}
\SetKwInOut{Output}{output}
\SetKwFunction{norm}{norm}
\SetKwFunction{whiten}{whiten}
\SetKwFunction{zeros}{zeros}

\Input{3d feature tensor $\bm{\cX}$,
pooling nhood size $w{\times}h$, stride $s$, and type $p$,
spatial weight generation function, $\Omega_s$,
channel weight generation function, $\Omega_c$,
initial norm type, $a$, and power scaling, $b$,
pre-trained whitening parameters $\bm{W}$,
final feature dimensionality $K^{\prime}$,
final norm type, $c$
}

\Output{$K^{\prime}$-dimensional aggregate feature vector $\bm{\cG} = \{ g_1, \ldots, g_K\}$}
\BlankLine

{${\widetilde{\bm{\cX}}}$ = pool($\bm{\cX}$;$w,h,s,p$) \tcp*{Initial local pooling}}
{ $\Omega_s(\tilde{\bm{\cX}})$ $\rightarrow$ $\spatialWeight$ $\forall$ $i,j$  \tcp*{Spatial weighting}  }
{ $\Omega_c(\tilde{\bm{\cX}})$ $\rightarrow$ $\channelWeight$ $\forall$ $k$ \tcp*{Channel weighting}  }
{$f_k = {\sum\limits_{i=1}^{W}}{\sum\limits_{j=1}^{H} \spatialWeight \channelWeight \cX_{kij}}$ $\forall$ $k$
}

{$\widehat{\bm{\cF}}$ = pnorm($\bm{\cF}$; $a$,$b$) \tcp*{Normalize and powerscale}}
{$\widetilde{\bm{\cF}}$ = PCA($\widehat{\bm{\cF}}$; $W$, $K^{\prime}$) \tcp*{dim. reduction and whitening}}
{$\bm{\cG}$ = norm($\widetilde{\bm{\cF}}$, $c$) \tcp*{Normalize again}}
\caption{Framework for Aggregation of Convolutional Features}
\label{alg:general_features}
\end{algorithm}

%% file: tex/feature_weighting.tex
\section{Feature Weighting Schemes}
\label{sec:weighting}

\noindent
In this section we present our non-parametric spatial and channel weighting for Steps 2 and 3 of the framework. We propose a spatial weighting derived from the spatial activations of the layer outputs themselves and a channel weighting derived from channel sparsity.

\input{tex/feature_weighting_spatial}

\input{tex/feature_weighting_channel_v2}


\subsection{Discussion}
\label{subsec:discussion}


\noindent

Using the framework described in Section~\ref{sec:framework}, we can explain different approaches in terms of their pooling, weighting and aggregation steps; we illustrate some interesting cases in Table~\ref{tab:spocsup}.
For example, approaches that aggregate the output of a max-pooling layer of the convolutional neural network are essentially performing max-pooling in Step 1.


In terms of novelty, it is noteworty to restate that the spatial weighting presented in Section~\ref{subsec:spam} corresponds to a well known principle, and approaches like~\cite{JeZi14, MuPe14, CiMV15} have addressed similar ideas. Our spatial weighting is notable as a simple and strong baseline. Together with the channel weighting, the \Crow features are able to deliver state-of-the-art results at practically the same computational cost as off-the-self features.
 
\head{Uniform weighting}.
If we  further uniformly set both spatial and channel weights and then perform sum-pooling per channel we end up with a simpler version of \Crow features, that we denote as \textit{uniform }\CroW or \uCrow. 

\head{Relation to SPoC~\cite{BaLe15} features}.
SPoC~\cite{BaLe15} can be described in terms of our framework as illustrated in Table~\ref{tab:spocsup}.
\Crow and SPoC features differ in their spatial pooling, spatial weighting, and channel weighting. For the first spatially-local pooling step, \Crow (and \uCrow) max-pool (we are essentially using the outputs of the last pooling layer of the deep convolutional network rather than the last convolutional one as in SpoC). SPoC uses a centering prior for spatial weighting to boost features that occur near the center of the image, whereas we propose a spatial weighting derived from the spatial activations of the layer outputs themselves. Lastly, SPoC uses a uniform channel weighting, whereas we propose a channel weighting derived from channel sparsity. We demonstrate improvements for each of these design choices in Section~\ref{sec:exp}.


%% file: tex/feature_weighting_spatial.tex
\subsection{Response Aggregation for Spatial Weighting}
\label{subsec:spam}

We propose a method to derive a spatial weighting based on the normalized total response across all channels. Let $\bm{\cS}^\prime \in \mathbb{R}^{(W \times H)}$ be the matrix of aggregated responses from all channels\textit{ per spatial location}, which we compute by \textit{summing} feature maps $\boldmatrixForChannel$: 


\begin{equation}
\bm{\cS}^\prime = \sum_{k} \boldmatrixForChannel.
\end{equation}

After normalization and power-scaling we get aggregated spatial response map $\bm{\cS}$, whose value at spatial location $(i,j)$ is given by:

\begin{equation}
\cS_{ij} =   \left (  \frac{S_{ij}^\prime}{ \left ( \sum_{m,n} {S_{mn}^\prime}^a \right ) ^{1/a} } \right ) ^{1/b},
\end{equation}

After computing the 2d spatial aggregation map $\bm{\cS}$ for feature $\bm{\cX}$, we can apply it independently on every channel, setting $\spatialWeight = \cS_{ij}$ and using $\spatialWeight$ as in Eqn~\ref{eqn:spatialweightsum}.




We experimented with different norms for normalizing the aggregate responses $\bm{\cS}^\prime$, \ie L1, L2, $\textit{inf}$, power normalization with $a = 0.5$~\cite{PeSM10}. We found that image search performance remains very high in all cases and the differences are very small, usually less than 0.01 in mAP. We therefore choose to use the L2 norm and $b=2$ for our spatial aggregation maps, before applying them to the features.

We visualize highly weighted spatial locations in Figure~\ref{fig:patches} with images from the \emph{Paris}~\cite{PCS+08} dataset. Our spatial weighting boosts features at locations with salient visual content and down weights non-salient locations. Notably, similar visual elements are boosted under our weighting despite large variation in lighting and perspective.

\begin{figure}[t]
\begin{subfigure}{0.56\textwidth}
\begin{tikzpicture}
\begin{axis}[
	height=6cm,
	xlabel={sparsity of$\lambda_{ij}$},
	ylabel={mean $\tilde{S_{ij}}$},
	legend pos=south west,
	minor y tick num=1,
    mark size=1
]
	\addplot[scatter/use mapped color={draw=cyan!70!black,fill=cyan!30!black},scatter,only marks] table[x=x, y=y, col sep=comma] {fig/true_plot.txt};
	\legend{}
\end{axis}
\end{tikzpicture}
\caption{}
\label{fig:sparsity2}
\end{subfigure}
\begin{subfigure}{0.43\textwidth}
\centering
\vspace{-.79cm}
\includegraphics[height=4.6cm]{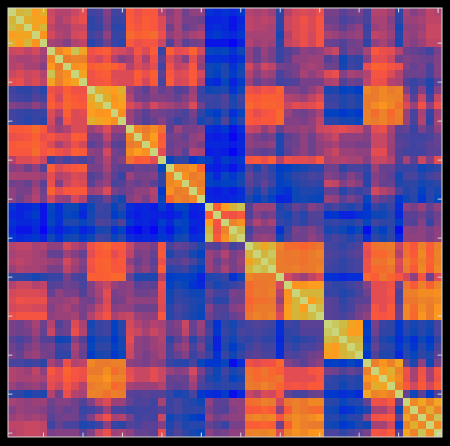}
\caption{}
\label{fig:sparsity_correlation}
\end{subfigure}
\caption{Fig.~\ref{fig:sparsity2}: Mean $\tilde{\cS_{ij}}$ plotted against channel sparsity at the corresponding location. Fig.~\ref{fig:sparsity_correlation}: The correlation of channel-wise sparsity for the $55$ images in the query-set of the \emph{Paris} dataset. Images are sorted by landmark class in both dimensions.}

\end{figure}

In Figure~\ref{fig:sparsity2} we show the relationship between our spatial weights $S_{ij}$ and the sparsity of the channel responses $\boldchannelsAtLocation$. We compute the spatial weight $S_{ij}$ of every location in the \emph{Paris} dataset and normalize each by the maximum spatial weight for the image in which it occurs, which we denote $\tilde{\cS_{ij}}$. The mean $\tilde{\cS_{ij}}$ for each level of channel sparsity at the corresponding location is plotted as cyan in Figure~\ref{fig:sparsity2}. 



It can be seen that our spatial weighting tends to boost locations for which multiple channels are active relative to other spatial locations of the same image. This suggests that our spatial weighting is a non-parametric and computationally cheap way to favor spatial locations for which features co-occur while also accounting for the strength of feature responses. We speculate that these locations are more discriminative as there are combinatorially more configurations at mid-ranges of sparsity.

\input{fig/patches}

%% file: fig/patches.tex
\begin{figure*}[t]

                
    \includegraphics[width=\linewidth]{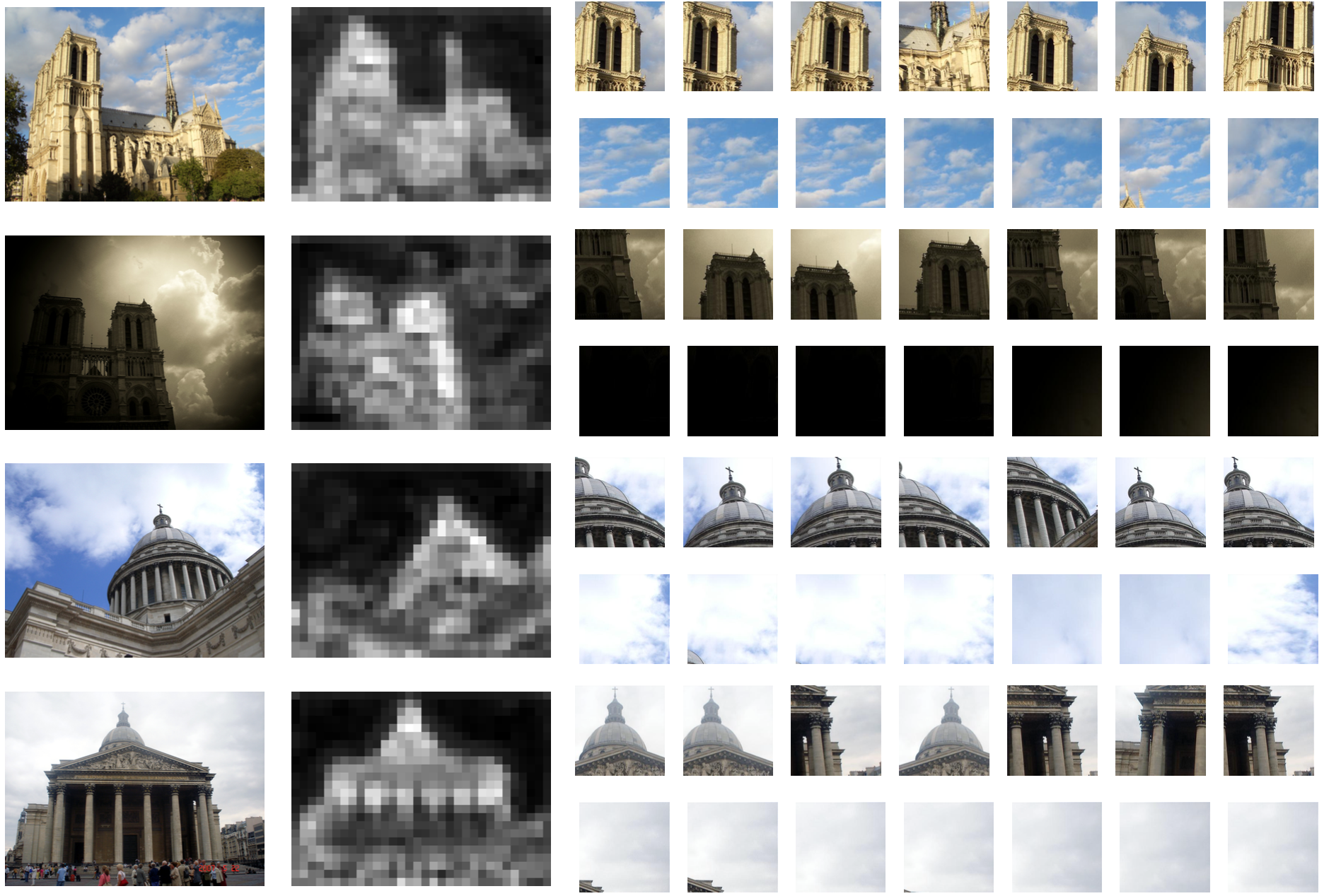}
    \caption{\small{Visualization of spatial weighting by aggregate response. On the left we show original images in the \emph{Paris} dataset along with their spatial weights. On the right we visualize the receptive fields of the 7 highest weighted locations and the 7 lowest weighted locations for each image. The top two images are of Notre Dame and the bottom two are of the Panth\'{e}on.}}
	\label{fig:patches}
\end{figure*}



 

%% file: tex/feature_weighting_channel_v2.tex
\subsection{Sparsity Sensitive Channel Weighting}
\label{subsec:idf}

We now propose a method to derive a channel weighting based on the sparsity of feature maps. We expect that similar images will have similar occurrence rates for a given feature. For each channel $k$ we find $\cQ_{k}$, the proportion of non-zero responses, and compute the \emph{per-channel sparsity}, $\Xi_k$, as:
\begin{equation}
\Xi_k = 1 - \cQ_{k},
\end{equation}
where  $\bm{\cQ} = \frac{1}{WH}\sum_{ij} \mathbbm{1}[\boldchannelsAtLocation > 0]$.
%
%
%
In Figure~\ref{fig:sparsity_correlation} we visualize the pair-wise correlation of the vectors of channel sparsities $\bm{\Xi} \in \mathbb{R}^{K}$ for images in the query-set of the \emph{Paris} dataset. The query-set for the \emph{Paris} dataset contains 55 images total, 5 images each for 11 classes of Paris landmarks. We order the images by class. It is apparent that channel sparsities $\bm{\Xi}$ are highly correlated for images of the same landmark and less correlated for images of different landmarks. It appears that the sparsity pattern of channels contains discriminative information.


Since we sum-pool features $\boldchannelsAtLocation$ over spatial locations when we derive our aggregated feature, channels with frequent feature occurrences are already strongly activated in the aggregate feature. However, infrequently occurring features could provide important signal if, for example, the feature consistently occurs though only a small number of times in images of the same class. Motivated by this insight, we devise a channel weighting scheme similar to the concept of \emph{inverse document frequency}. That is, we boost the contribution of rare features in the overall response by using the per-channel weight, $\cI_{k}$, defined as:

\begin{equation}
\cI_{k} = \log \left(\frac{ K\epsilon + \sum_{h} \cQ_{h}}{\epsilon + \cQ_{k} }\right),
\end{equation}

where $\epsilon$ is a small constant added for numerical stability.

Our sparsity sensitive channel weighting is also related to and motivated by the notion of intra-image visual burstiness~\cite{JeDS09}. Channels with low sparsity correspond to filters that give non-zero responses in many image regions. This implies some spatially recurring visual elements in the image, that were shown to negatively affect matching~\cite{JeDS09}. Although we don't go as far as~\cite{JeZi14} and try to learn a ``democratic'' matching kernel, our sparsity sensitive weights do down-weight channels of such bursty convolutional filters.

To provide further insight into the effect of our sparsity-sensitive channel weights (SSW), we visualize the receptive fields of active locations in channels that our weights boost.

\begin{figure}[t]
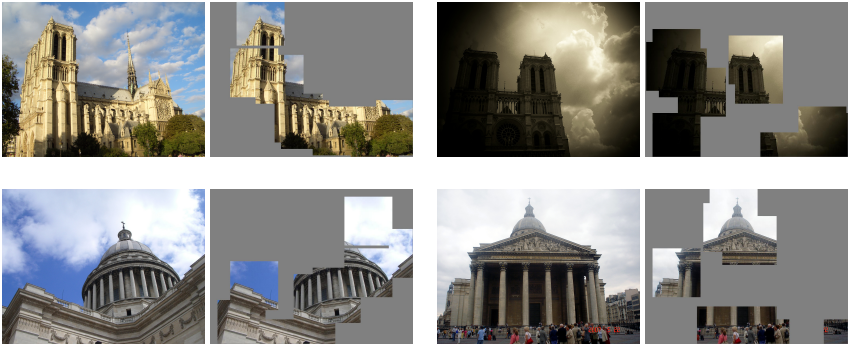

	\centering
    	\foreach \i in {paris_notredame_000035, paris_notredame_000107} {
		\includegraphics[width=2.7cm]{fig/\i.jpg} 
		\includegraphics[width=2.7cm]{fig/\i_top_patches_all.jpg}
		\hspace{.1cm}  
		\vspace{.2cm}
    }
    \\
    \foreach \i in {paris_pantheon_000014, paris_pantheon_000013} {
		\includegraphics[width=2.7cm]{fig/\i.jpg} 
		\includegraphics[width=2.7cm]{fig/\i_top_patches_all.jpg} \hspace{.1cm}  
		\vspace{.2cm}
    }
\caption{Regions corresponding to locations that contribute (are non-zero) to the $10$ channels with the highest sparsity-sensitive weights for the four images of Figure~\ref{fig:patches}.}
\label{fig:idfviz}
\end{figure}

In Figure~\ref{fig:idfviz} we show all receptive fields that are non-zero in (one or more) of the channels with the highest sparsity-sensitive channel weights. As values from these channels are increased before aggregation, our approach gives more weight to CNN outputs that correspond to the image regions shown on the right.

\begin{table}[t]
\setlength{\tabcolsep}{12pt}
\small {
	\begin{tabularx}{\columnwidth}{X c c c}
  		Step & SPoC~\cite{BaLe15} & \uCrow & \Crow \\
  		\midrule
  		1: local pooling & none & max & max \\
  		2: spatial weighting & centering prior & uniform & \sCrow \\
  		3: channel weighting & uniform & uniform & \cCrow \\
        4: aggregation & sum & sum & sum
	\end{tabularx}
}
\vspace{.15cm}
\caption{The pooling and weighting steps for three instantiations of our aggregation framework, \ie the proposed \Crow, the simplified \uCrow and SPoC~\cite{BaLe15}. SW refers to the spatial weighting presented in~\ref{subsec:spam}, while SSW to the sparsity sensitive channel weighting presented in Section~\ref{subsec:idf}.}
\label{tab:spocsup}
\end{table}




%% file: tex/experiments.tex
\section{Experiments}
\label{sec:exp}

\noindent

\input{tex/experiments_protocol}

\input{tex/experiments_prelim}

\input{tex/experiments_search}

%% file: tex/experiments_protocol.tex
\subsection{Evaluation Protocol}
\label{subsec:eval}

\subsubsection{Datasets}

We experiment on four publicly available datasets. For image search we report results on \emph{Oxford}~\cite{PCI+07} and \emph{Paris}~\cite{PCS+08}, further combining them with the \emph{Oxford100k}~\cite{PCI+07} dataset as distractors. We also present results on the \emph{Holidays}~\cite{JeDS08} dataset.
For Oxford we used the common protocol as in all other methods reported, \emph{i.e.} the cropped queries. We regard the cropped region as input for the CNN and extract features. For Holidays we use the ``upright'' version of the images.

\subsubsection{Evaluation Metrics}

For image search experiments on \emph{Oxford}, \emph{Paris} and \emph{Holidays} we measure mean average precision (mAP) over all queries. We use the evaluation code provided by the authors.
For deep neural networks we use Caffe\footnote{\url{http://caffe.berkeleyvision.org/}}~\cite{YSD+14} and the publicly available pre-trained VGG16 model~\cite{SiZi14}. As usual with Caffe, we zero-center the input image by \textit{mean pixel subtraction}.
In all cases, table rows including citations present results reported in the cited papers.

\subsubsection{Query Expansion}

One can trivially use simple query expansion techniques~\cite{CPS+07} with \Crow features. Given the ranked list of database images by ascending distance to the query, we sum the aggregated feature vectors of the top $M$ results, L2-normalize and re-query once again. Despite its simplicity, we show that this consistently improves performance, although it does come at the cost of one extra query.

%% file: tex/experiments_prelim.tex
\subsection{Preliminary Experiments}
\label{subsec:exp_prelim}

\head{Image size and layer selection}. In Figure~\ref{fig:layers} we investigate the performance of \Sup features when aggregating responses from different layers of the network.

Our \uCrow features are in essence similar to the very recently proposed SPoC features of~\cite{BaLe15}, but have some different design choices that make them more generic and powerful. Firstly, SPoC features are derived from the VGG19 model while our \SuP features are derived from the VGG16 model; in this section we show that our \SuP features performs much better even if we are using a smaller deep network. Secondly, we do not resize the input image to $586 \times 586$ as in~\cite{BaLe15} and instead keep it at its original size. SpoC is therefore comparable to the dotted cyan line in Figure~\ref{fig:layers}.

Choosing the last pooling and convolutional layers of the network significantly improves performance over the fourth, especially as the final dimension decreases. Moreover, the \texttt{pool5} layer consistently outperforms \texttt{conv5-3}, showing that max pooling in Step 1 is indeed beneficial. 

Regarding image size, we see that keeping the original size of the images is another factor that contributes to higher performance.

\begin{figure}[t]
\centering
\begin{subfigure}{0.49\textwidth}
\begin{tikzpicture}
\begin{semilogxaxis}[%
	height=5cm,
	xlabel={dimensions},
	ylabel={mAP},
	legend pos=south east,
    ymax = 0.8,
    ymin = 0.65,
	minor y tick num=1,
    xtick={128,256,512},    
    xticklabels={128,256,512}
]
	\pgfplotstableread{
		d    pool5 conv5 conv4 pool4 fspool5 fsconv5 c4 p4   
		512  0.767  0.748 0.740 0.741 0.737 0.723 nan nan
		256  0.759  0.744 0.699 0.695 0.734 0.721 0.699 0.695
		128  0.729  0.702 nan nan 0.708 0.694 0.58 0.58

	}{\map}
	\addplot[black,mmark]   table[y=pool5]  \map; \leg{\tiny{\texttt{pool5}}}
	\addplot[cyan,mmark]   table[y=conv5]     \map; \leg{\tiny{\texttt{conv5}}}
	\addplot[magenta,mmark]    table[y=pool4]    \map; \leg{\tiny{\texttt{pool4}}}
	\addplot[yellow!70!black,mmark]     table[y=conv4]     \map; \leg{\tiny{\texttt{conv4}}    }

\addplot[black!90!black,mmark, dash]     table[y=fspool5]     \map; \leg{\tiny{\texttt{pool5(*)}}}
	\addplot[cyan!90!black,mmark, dash]     table[y=fsconv5]     \map; \leg{\tiny{\texttt{conv5(*)}}}
	\addplot[magenta!40!white,mmark]    table[y=p4]    \map; 
	\addplot[yellow!40!white,mmark]     table[y=c4]     \map;   

\end{semilogxaxis}
\end{tikzpicture}
\caption{}
\label{fig:layers}
\end{subfigure}
\begin{subfigure}{0.49\textwidth}
\begin{tikzpicture}
\begin{semilogxaxis}[%
	height=5cm,
	xlabel={dimensions},
	ylabel={mAP},
	legend pos=south east,
	minor y tick num=1,
    xtick={128,256,512},    
    xticklabels={128,256,512}
]
	\pgfplotstableread{
		d    o 		oscm   oscmidf oidf oscmidfqe
		512  0.767  0.780  0.796  0.787 0.855
		256  0.759  0.772  0.779  0.775 0.850
		128  0.729  0.744  0.746  0.734 0.827
	}{\map}
	\addplot[black,mmark]   table[y=oscmidf]  \map; \leg{\supp}
	\addplot[cyan,mmark]   table[y=oscm]     \map; \leg{\Sup+\SCM}
	\addplot[magenta,mmark]    table[y=oidf]    \map; \leg{\Sup+\IDF}
	\addplot[green!80!black,mmark]     table[y=o]     \map; \leg{\Sup}
\end{semilogxaxis}
\end{tikzpicture}
\caption{}
\label{fig:paris_dim}
\end{subfigure}

\caption{Fig. \ref{fig:layers}: Mean average precision on Paris. Different lines denote \uCrow features from the corresponding layers of VGG16; \texttt{conv4} (\texttt{conv5}) corresponds to \texttt{conv4\_3} (\texttt{conv5\_3}). Solid lines denote that the original image size is kept, while for dashed lines the images were resized to $586 \times 586$ as in~\cite{BaLe15}.  Both \texttt{conv4} and \texttt{pool4} layers have very poor performance in low dimensions, with $0.58$ mAP for $d=128$. SPoC features~\cite{BaLe15} correspond to the dotted cyan line. Fig. \ref{fig:paris_dim}: Mean average precision on \emph{Paris} when varying the dimensionality of the final features.}

\end{figure}


\head{Effect of the final feature dimensionality}.
In Figure~\ref{fig:paris_dim} we present mAP on \emph{Paris} when varying the dimensionality of the final features. We present results for all weighting combinations of the proposed approach. \uCrow refers to uniform or no weighting. \uCrow+\sCrow refers to using the only the spatial weighting of Section~\ref{subsec:spam} on top of \uCrow, \uCrow+\cCrow to using the sparsity sensitive channel weighting of Section~\ref{subsec:idf} on top of \uCrow, while \Crow refers to our complete approach with both weighting schemes. As we see, the \uCrow+\cCrow combination is affected more by dimensionality reduction than the rest. This can be interpreted as an effect of the subsequent dimensionality reduction. When calculating the sparsity sensitive weights all dimensions are taken into account, however, in the final reduced vector many of those were discarded.

\head{Notes on max-pooling}.
In preliminary experiments we also tested  max-pooling instead of sum pooling for feature aggregation. Consistently with~\cite{BaLe15} we found it to be always inferior to sum-pooling when whitening was used. Interestingly, max pooling performs better than sum-pooling in the non-whitened space, but mAP without whitening the features is much inferior (sometimes more than $10\%$ less) in all datasets tested.

\head{Whitening}. We learn the whitening parameters from a separate set of images. In Table~\ref{tab:whiten} we present results on the \emph{Paris} dataset when using 3 other datasets for whitening: the semantically related \emph{Oxford} dataset, the \emph{Holidays} dataset and the larger \emph{Oxford100k} dataset. As we reduce the dimensionality, we see overfitting effects for the case where we learn on \emph{Oxford} and this comes as no surprise: as dimensions are reduced, more dimensions that are selective for buildings are kept when we learn the reduction parameters on a semantically similar dataset like \emph{Oxford}.

To be directly comparable with related works, we learn the whitening parameters on \emph{Oxford} when testing on \emph{Paris} and vice versa, as accustomed. We use the \emph{Oxford100k} dataset for whitening on the \emph{Holidays}.

\alert{
\begin{table}[t]
\setlength{\tabcolsep}{12pt}
\centering

\begin{tabularx}{.9\columnwidth}{c X c c c}
 $d$					 &		 & Oxford & Holidays & Oxford100k \\
\multirow{2}{*}{512} & \uCrow  & 0.786 & 0.752 & 0.803\\
                     & \Crow & 0.797 & 0.792 & \textbf{0.810}\\
\midrule
\multirow{2}{*}{256} & \uCrow  & 0.739 & 0.728 & 0.732\\
                     & \Crow & \textbf{0.765} & 0.784 & 0.762 \\

\end{tabularx}

\caption{Mean average precision on Paris when learning the whitening parameters on Oxford, Holidays and Oxford100k for different values of $d$.}
\label{tab:whiten}
\end{table}
}

%% file: tex/experiments_search.tex
\begin{table}[th!]
\setlength{\tabcolsep}{5pt}
\centering
\small{
\begin{tabularx}{\columnwidth}{X c  c c  c c  c}
Method & d & Paris & +Oxf100k & Oxford & +Oxf100k & Holidays\\
\midrule
Tr. Embedding~\cite{JeZi14} & 1024 & --- & --- & 0.560 & 0.502 & 0.720 \\
\midrule
Tr. Embedding~\cite{JeZi14} & 512 & --- & --- & --- & --- & 0.700 \\
Gong~\etal~\cite{BSCL14} 	& 512 & --- & --- & --- & --- & 0.783  \\
Neural Codes~\cite{BSCL14} 	& 512 & --- & --- & 0.435  &  0.392 & --- \\
R-MAC~\cite{ToSJ15} & 512 	& \textbf{0.830} & \textbf{0.757} & 0.669 & 0.616 & --- \\
\uCrow 						& 512 & 0.786 & 0.710 & 0.697 & 0.641  & 0.839 \\
\Crow 						& 512 & 0.797 & 0.722 & \textbf{0.708} & \textbf{0.653} & \textbf{0.851} \\
\midrule
Tr. Embedding~\cite{JeZi14} & 256 & --- & --- & --- & --- & 0.657  \\
Neural Codes~\cite{BSCL14} & 256  & --- & --- & 0.435  &  0.392 & 0.749 \\
Razavian \etal~\cite{RSMC14} & 256 & 0.670 & ---  & 0.533 & 0.489 & 0.716 \\
SPoC~\cite{BaLe15} & 256 & --- & --- & 0.531 & 0.501  & 0.802 \\
R-MAC~\cite{ToSJ15} & 256 & 0.729 & 0.601  & 0.561 & 0.470  & ---\\
\uCrow & 256 & 0.739 & 0.658 & 0.667 & 0.612 & 0.815\\
\Crow & 256 & \textbf{0.765} & \textbf{0.691} &  \textbf{0.684} & \textbf{0.637}  & \textbf{0.851}  \\
\midrule
Tr. Embedding~\cite{JeZi14} & 128 & --- & --- & 0.433 & 0.353 & --- \\
Neural Codes~\cite{BSCL14} & 128  & --- & --- & 0.433  & 0.384 & --- \\
\uCrow & 128 & 0.699 & 0.610 & 0.625 & 0.559 & --- \\
\Crow & 128 & \textbf{0.746} & \textbf{0.670} & \textbf{0.641} & \textbf{0.590} & \textbf{0.828}\\
\midrule
\Crow + QE & 128 & 0.793 & 0.728 & 0.670 & 0.641 & ---\\
\Crow + QE & 256 & 0.815 & 0.753 & 0.718 & 0.676 & --- \\
\Crow + QE & 512 & \textbf{0.848} & \textbf{0.794} & \textbf{0.749} & \textbf{0.706} & ---\\
\midrule
Tolias~\etal~\cite{ToJA15} & --- & 0.770 & --- & 0.804 & 0.750 & ---\\
Total Recall II~\cite{CMPM11} & --- & 0.805 & 0.710 & 0.827 &  0.767 & ---\\
Mikulik~\etal~\cite{MPCM10} & --- & 0.824 &  0.773  & 0.849 &  0.795 & ---\\
\vspace{.05cm}
\end{tabularx}
}
\caption{Mean average precision on \emph{Paris}, \emph{Oxford} and \emph{Holidays} against the state-of-the-art for different values of $d$. QE denotes query expansion with the top $M=10$ results. The fourth (sixth) column presents results when augmenting the \emph{Paris} (\emph{Oxford}) dataset with the 100k distractors from \emph{Oxford100k}. Results in the lowest set of rows correspond to methods with local features, followed by spatial verification.}
\label{tab:search}
\vspace{-10pt}
\end{table}

\subsection{Image Search}
\label{subsec:exp_search}

In Table~\ref{tab:search} we present comparisons of our approach with the state-of-the-art in image search on \emph{Paris}, \emph{Oxford} and \emph{Holidays}. Both \Sup and \supp consistently outperform all other aggregation methods for different representation sizes, apart from R-MAC~\cite{ToSJ15}, which exhibits very high performance for Paris in 512 dimensions. 

\uCrow is a very strong baseline that gives state-of-the-art performance by itself. It therefore makes sense that improvement over \uCrow is hard to get. \Crow improves performance in all cases, with the gain increasing as the dimensionality of the final features decreases. 
For comparison, if we apply our weighting (instead of the centering prior) to SPoC features, the gain on \emph{Paris} is around $3.9\%$ and $4.6\%$ for 256 and 512 dimensions respectively.

\begin{figure}[t]
	\centering
		\includegraphics[width=.8\linewidth]{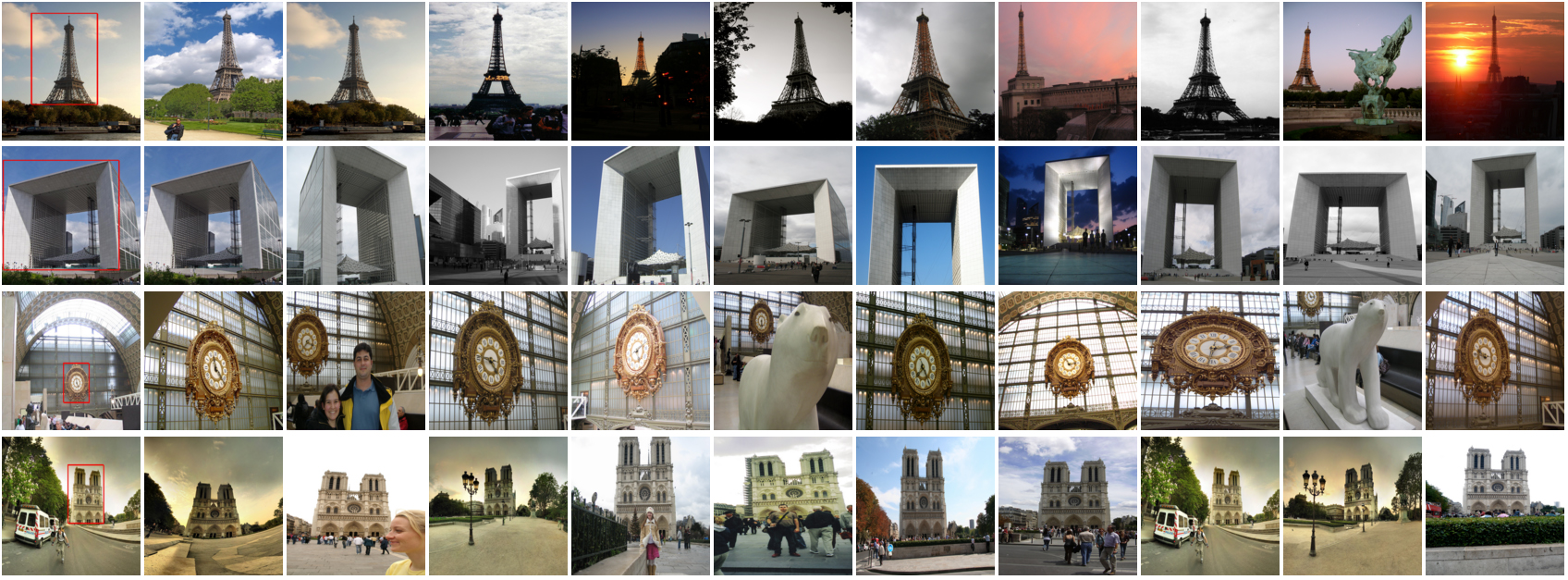} 
\caption{Sample search results using \Crow features compressed to just $d=32$ dimensions. The query image is shown at the leftmost side with the query bounding box marked in a red rectangle.}
\label{fig:resgood}
\end{figure}

In Figure~\ref{fig:resgood} we present some interesting results \emph{using just $d=32$ dimensional features}. They demonstrate the invariance of \Crow features to viewpoint and lighting variations even after heavy compression.

When further combining our approach with query expansion, we get even better results that compare to (or surpass on \emph{Paris}) far more sophisticated approaches like~\cite{MPCM10,CMPM11,ToJA15} that are based on local features and include spatial verification steps.

In Figure~\ref{fig:res} we show the top-10 results for \textit{all} 55 queries on the paris dataset using the uncompressed \Crow features ($d=512$). \textbf{We only have 3 false results in total} for precision@10. This illuminates why query expansion is so effective: the top ranked results are already of high quality.

Although our approach is consistently better, the performance gap between \Crow and the state-of-the-art is smaller in \emph{Holidays}, where it outperforms the best competing method by about $4.9\%$ and $1.2\%$ for $d=256,512$, respectively.

%% file: tex/conclusions.tex
\section{Conclusions}
\label{sec:conclusions}

In this paper we outline a generalized framework for aggregated deep convolutional features with cross-dimensional weighting which encompasses recent related works such as~\cite{BaLe15}. We propose simple, non-parametric weighting schemes for spatial- and channel-wise weighting and provide insights for their behavior by visualizing and studying the distributional properties of the layer output responses. Using this approach, we report results that outperform the state-of-the-art in popular image search benchmarks. 

The \Crow features are one instantiation of our generic aggregation framework. Still, it gives the current state-of-the-art results in image retrieval with minimal overhead and has intuitive qualities that offer insights on the nature of convolutional layer features.

Our aggregation framework is a valuable scaffold within which to discuss and explore new weighting schemes. The framework gave us a clear way to investigate channel and spatial weights independently. Learning weights for a particular task is a promising future direction. Likewise, with sufficient ground truth data, it is possible to fine-tune the entire end-to-end process within our proposed framework using, say, a rank-based loss as in~\cite{RaTC16, GARL16}, together with \textit{attentional mechanisms} and spatial deformations~\cite{JSZK15}. 


\begin{figure}[th!]
	\centering
		\includegraphics[width=.45\linewidth]{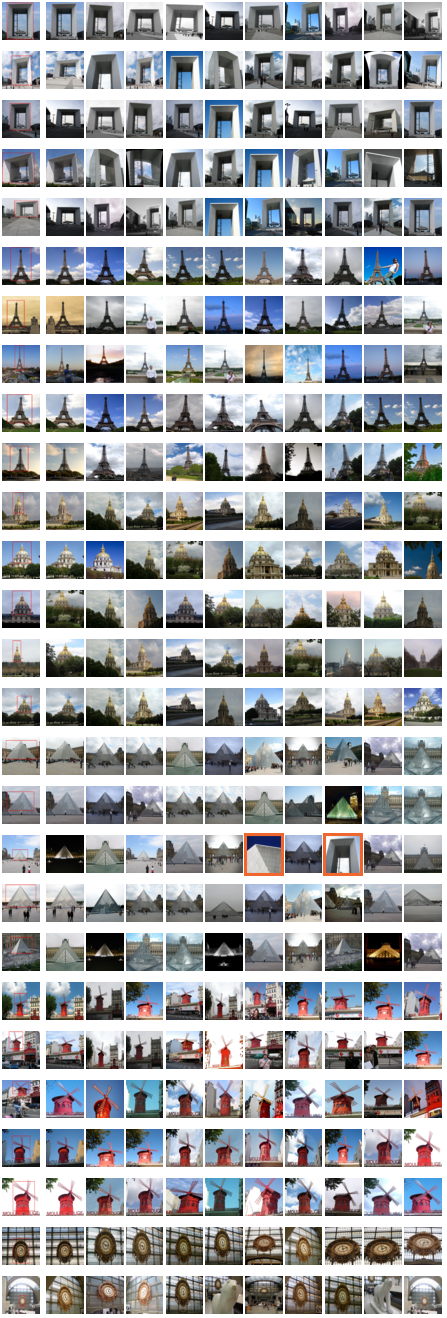} 
		\includegraphics[width=.45\linewidth]{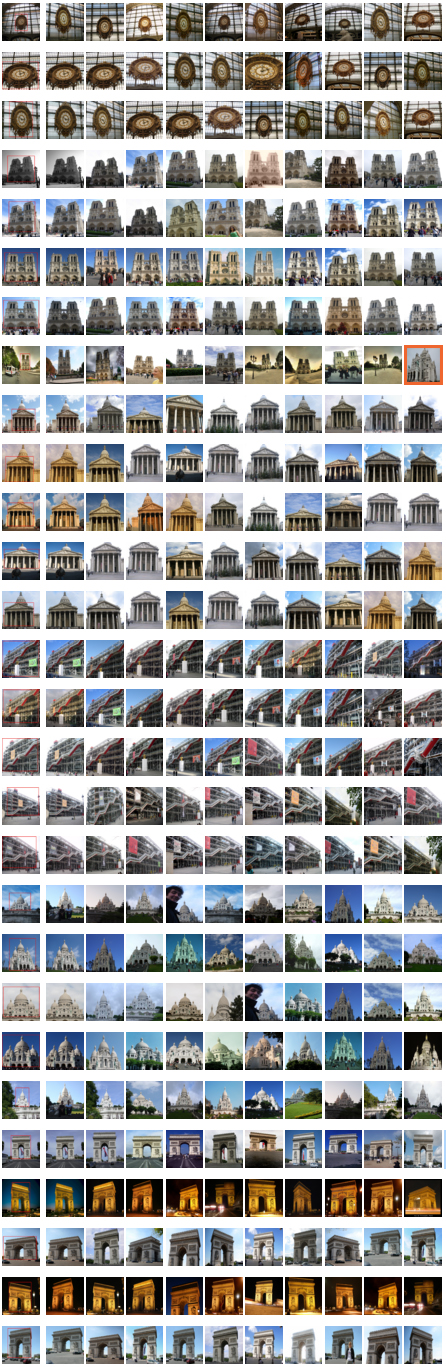} 
        
\caption{Top-10 results returned for all 55 queries of the Paris dataset, using the 512-dimensional \Crow features (and no query expansion). The query image is shown on the leftmost place, with the query bounding box marked with a red rectangle. Our features produce just 3 false results in total, which are marked with an orange border.}
\label{fig:res}
\end{figure}
\clearpage